\documentclass[11pt,a4paper]{article}
\usepackage{authblk}
\usepackage[hyperref]{naaclhlt2019}
\usepackage{times}
\usepackage{latexsym}

\usepackage{caption} 
\usepackage{array}
\usepackage{url}
\usepackage{algorithm, algorithmic}
\newcommand{\INDSTATE}[1][1]{\STATE\hspace{#1\algorithmicindent}}
\usepackage{graphicx}
\usepackage{amsmath}
\usepackage{booktabs}

\newcolumntype{L}{>{\centering\arraybackslash}m{1.5cm}}
\aclfinalcopy 

\title{Generative Adversarial Networks for text using word2vec intermediaries}

\author[1, 2]{Akshay Budhkar}
\author[1]{Krishnapriya Vishnubhotla}
\author[1, 2]{Safwan Hossain}
\author[1, 2, 3]{Frank Rudzicz}
\affil[1]{Department of Computer Science, University of Toronto}
\affil[2]{Vector Institute}
\affil[3]{St Michael's Hospital}

\date{}

\begin{document}
\maketitle
\begin{abstract}
Generative adversarial networks (GANs) have shown considerable success, especially in the realistic generation of images. In this work, we apply similar techniques for the generation of text. We propose a novel approach to handle the discrete nature of text, during training, using word embeddings. Our method is agnostic to vocabulary size and achieves  competitive results relative to methods with various discrete gradient estimators.
\end{abstract}

\section{Introduction}
Natural Language Generation (NLG) is often regarded as one of the most challenging tasks in computation \citep{murty1987some}. It involves training a model to do language generation for a series of abstract concepts, represented either in some logical form or as a knowledge base. Goodfellow introduced generative adversarial networks (GANs) \citep{goodfellow2014generative} as a method of generating synthetic, continuous data with realistic attributes. The model includes a discriminator network ($D$), responsible for distinguishing between the real and the generated samples, and a generator network ($G$), responsible for generating realistic samples with the goal of fooling the $D$. This setup leads to a minimax game where we maximize $V$ with respect to $D$, and minimize it with respect to $G$. The ideal optimal solution is the complete replication of the real distributions of data by the generated distribution.

GANs, in this original setup, often suffer from the problem of mode collapse - where the $G$ manages to find a few modes of data that resemble real data, using them consistently to fool the $D$. Workarounds for this include updating the loss function to incorporate an element of multi-diversity. An optimal $D$ would provide $G$ with the information to improve, however, if at the current stage of training it is not doing that yet, the gradient of $G$ vanishes. Additionally, with this loss function, there is no correlation between the metric and the generation quality, and the most common workaround is to generate targets across epochs and then measure the generation quality, which can be an expensive process.

W-GAN \citep{arjovsky2017wasserstein} rectifies these issues with its updated loss. Wasserstein distance is the minimum cost of transporting \textit{mass} in converting data from distribution $P_r$ to $P_g$. This loss \emph{forces} the GAN to perform in a min-max, rather than a max-min, a desirable behavior as stated in \citep{goodfellow2016nips}, potentially mitigating mode-collapse problems. The loss function is given by:

\begin{equation}
\label{eq:critic}
L_{critic} = \min_{G} \max_{D \in \mathcal{D}}  E_{x\sim p_r(x)}[ D(x)] - E_{\tilde{x}\sim p_g(x)}[ D(\tilde{x})]
\end{equation}

where $\mathcal{D}$ is the set of 1-Lipschitz functions and $P_g$ is the model distribution implicitly defined by $\tilde{x} = G(z)$, $z \sim p(z)$. A differentiable function
is 1-Lipschtiz \emph{iff} it has gradients with norm at most 1 everywhere. Under an optimal $D$ minimizing the value function with respect to the generator parameters minimizes the $\mathcal{W}(p_r, p_g)$, where $\mathcal{W}$ is the Wasserstein distance, as discussed in \citep{vallender1974calculation}. To enforce the Lipschitz constraint, the authors propose clipping the weights of the gradient within a compact space $[-c, c]$. 

\citep{gulrajani2017improved} show that even though this setup leads to more stable training compared to the original GAN loss function, the architecture suffers from exploding and vanishing gradient problems. They introduce the concept of \emph{gradient penalty} as an alternative way to enforce the Lipschitz constraint, by penalizing the gradient norm directly in the loss. The loss function is given by:

\begin{equation}
\label{eq:gp}
L = L_{critic} + \lambda E_{\hat{x}\sim{p(\hat{x})}} [( || \nabla_{\hat{x}}D(\hat{x})||_{2} - 1)^2]
\end{equation}

where $\hat{x}$ are random samples drawn from $P_x$, and $L_{critic}$ is the loss defined in Equation \ref{eq:critic}.

Empirical results of GANs over the past year or so have been impressive. GANs have gotten state-of-the-art image-generation results on datasets like ImageNet \citep{brock2018large} and LSUN \citep{radford2015unsupervised}.  Such GANs are fully differentiable and allow for back-propagation of gradients from $D$ through the samples generated by G. However, if the data is discrete, as, in the case of text, the gradient cannot be propagated back from $D$ to G, without some approximation. Workarounds to this problem include techniques from reinforcement learning (RL), such as policy gradients to choose a discrete entity and reparameterization to represent the discrete quantity in terms of an approximated continuous function \citep{williams1992simple, jang2016categorical}.

\subsection{Techniques for GANs for text}

SeqGAN \citep{yu2017seqgan} uses policy gradient techniques from RL to approximate gradient from discrete $G$ outputs, and applied MC rollouts during training to obtain a loss signal for each word in the corpus. MaliGAN \citep{che2017maximum} rescales the reward to control for the vanishing gradient problem faced by SeqGAN. RankGAN \citep{lin2017adversarial} replaces $D$ with an \textit{adversarial ranker} and minimizes pair-wise ranking loss to get better convergence, however, is more expensive than other methods due to the extra sampling from the original data. \cite{kusner2016gans} used the Gumbel-softmax approximation of the discrete one-hot encoded output of the G, and showed that the model learns rules of a context-free grammar from training samples. \cite{rajeswar2017adversarial}, the state of the art in 2017, forced the GAN to operate on continuous quantities by approximating the one-hot output tokens with a softmax distribution layer at the end of the $G$ network. 

MaskGAN \citep{fedus2018maskgan} uses policy gradient with REINFORCE estimator \citep{williams1992simple} to train the model to predict a word based on its context, and show that for the specific blank-filling task, their model outperforms maximum likelihood model using the perplexity metric. LeakGAN \citep{guo2018long} allows for long sentence generation by \textit{leaking} high-level information from $D$ to G, and generates a latent representation from the features of the already generated words, to aid in the next word generation. TextGAN \citep{zhang2017adversarial} adds an element of diversity to the original GAN loss by employing the Maximum Mean Discrepancy objective to alleviate mode collapse.

In the latter half of 2018, \cite{zhu2018texygen} introduced \textit{Texygen}, a benchmarking platform for natural language generation, while introducing standard metrics apt for this task. \cite{lu2018neural} surveys all these new methods along with other baselines, and documents model performance on standard corpus like  EMNLP2017 WMT News\footnote{\url{http://www.statmt.org/wmt17/}} and Image COCO\footnote{\url{http://cocodataset.org/}}.

\section{Motivation}
\subsection{Problems with the Softmax Function}

\label{sec:softmax}
The final layer of nearly all existing language generation models is the softmax function. It is usually the slowest to compute, leaves a large memory footprint and can lead to significant speedups if replaced by approximate continuous outputs \citep{kumar2018mises}. Given this bottleneck, models usually limit the vocabulary size to a few thousand and use an unknown token (\textit{unk}) for the rare words. Any change in the allowed vocabulary size also means that the researcher needs to modify the existing model architecture. 

Our work breaks this bottleneck by having our $G$ produce a sequence (or stack) of continuous distributed word vectors, with $n$ dimensions, where $n << V$ and $V$ is the vocabulary size. The expectation is that the model will output words in a semantic space, that is produced words would either be correct or close synonyms \citep{mikolov2013distributed,kumar2018mises}, while having a smaller memory footprint and faster training and inference procedures.

\subsection{GAN2vec}

In this work, we propose GAN2vec - GANs that generate real-valued word2vec-like vectors (as opposed to discrete one-hot encoded outputs). While this work mainly focuses specifically on word2vec-based representation, it can be easily extended to other embedding techniques like GloVe and fastText.

Expecting a neural network to generate text is, intuitively, expecting it to learn all the nuances of natural language, including the rules of grammar, context, coherent sentences, and so on. Word2vec has shown to capture parts of these subtleties by capturing the inherent semantic meaning of the words, and this is shown by the empirical results in the original paper \citep{mikolov2013distributed} and with theoretical justifications by \citep{ethayarajh2018towards}. GAN2vec breaks the problem of generation down into two steps, the first is the word2vec mapping, with the following network expected to address the other aspects of sentence generation. It also allows the model designers to swap out word2vec for a different type of word representation that is best suited for the specific language task at hand.

As a manifestation of the “similar-context” words getting grouped in word embedding space - we expect GAN2vec to have synonymic variety in the generation of sentences. Generating real-valued word vectors also allows the $G$ architecture to be vocabulary-agnostic, as modifying the training data would involve just re-training the word embedding with more data. While this would involve re-training the weights of the GAN network, the initial architectural choices could remain consistent through this process. Finally, as discussed in Section \ref{sec:softmax}, we expect a speed-up and smaller memory footprint by adapting this approach.

All the significant advances in the adaptation of GANs since its introduction in 2016, has been focused in the field of images. We have got to the point, where sometimes GAN architectures have managed to generate images even \textit{better} than real images, as in the case of BigGAN \citep{brock2018large}. While there have been breakthroughs in working with text too, the rate of improvement is no-where close to the success we have had with images. GAN2vec attempts to bridge this gap by providing a framework to swap out image representations with word2vec representations.

\section{The Architecture}

\begin{figure*}[tbp]
\centering
  \includegraphics[width=1.5\columnwidth]{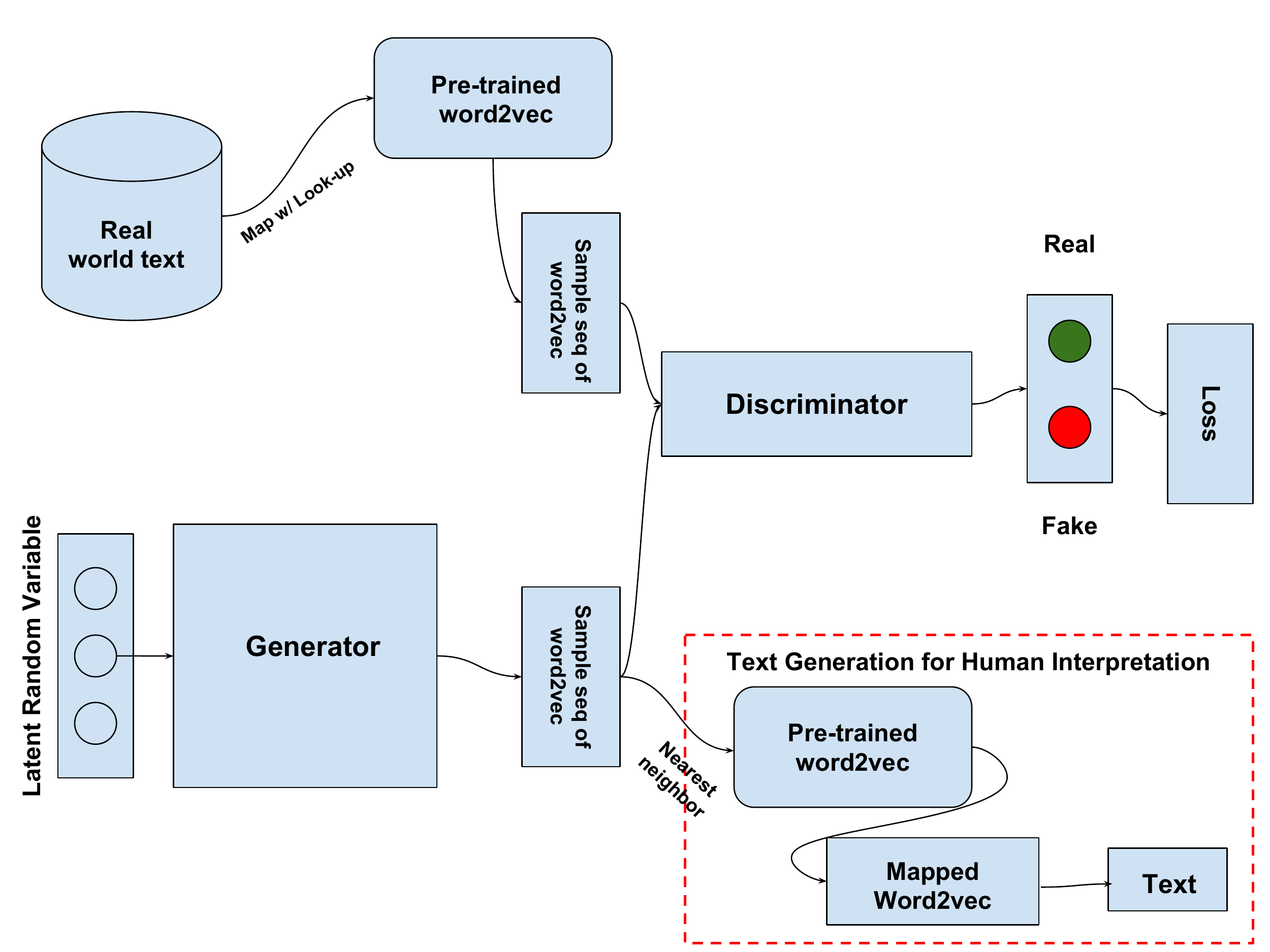}
  \caption{Structure of the GAN2vec model. Random normal noise is given as input to the generator network $G$. The discriminator network $D$ is responsible for determining whether a sample originated from $G$ or from the training set. At inference time, we use a nearest-neighbor approach to convert the output from $G$ into human-readable text.}
  \label{fig:gans}
\end{figure*}

Random normal noise is used as an input to the $G$ which generates a sequence of word2vec vectors. We train the word2vec model on a real text corpus and generate a stack word vector sequences from the model. The generated and the real samples are then sent to $D$, to identify as real or synthetic. The generated word vectors are converted to text at regular intervals during training and during inference for human interpretation. A nearest-neighbor approach based on cosine similarity is used to find the closest word to the generated embedding in the vector space.

\section{The Algorithm}
The complete GAN2vec flow is presented in Algorithm~\ref{al1}. 

\begin{algorithm}[ht]
\caption{GAN2vec Framework}
\begin{algorithmic}[1]
\label{al1}
\STATE Train a word2vec model, $e$, on the train corpus
\STATE Transform text to a stack of word2vec vectors using $e$
\STATE Pre-train $D$ for \textit{t} iterations on real data
\STATE \textbf{for} \textit{k} iterations \textbf{do}
\INDSTATE Send minibatch of real data to \emph{D}
\INDSTATE $G(z)$ = Sample random normal \emph{z} and feed to \emph{G}
\INDSTATE Send minibatch of \emph{G} generated data, $G(z)$, to \emph{D}
\INDSTATE Update \emph{D} using gradient descent
\INDSTATE Update \emph{G} using gradient ascent
\STATE end \textbf{for}
\STATE $G(z)$ = Sample random normal \emph{z} and feed to \emph{G}
\STATE $w_{generated} = \underset{w}{argmin}\{d(\hat{e}, e(w))\}$, for every $\hat{e}$ in $G(z)$ and every $w$ in the corpus
\end{algorithmic}
\end{algorithm}

\subsection{Conditional GAN2vec}
We modify GAN2vec to measure the adaptability of GAN2vec to conditions provided \textit{a priori}, as seen in \citep{mirza2014conditional}. This change can include many kinds of conditions like positive/negative, question/statement or dementia/controls, allowing for the ability to analyze examples from various classes on the fly during inference. Both the $G$ (and $D$) architectures get passed the condition at hand as an input, and the goal of $G$ now is to generate a realistic sample given the specific condition.

\section{Environmental Setup}

All the experiments are run using Pytorch \citep{paszke2017pytorch}. Word2vec training is done using the \textit{gensim} library \citep{rehurek_lrec}. Unless specified otherwise, we use the default parameters for all the components of these libraries, and all our models are trained for \textbf{100} epochs. The word embedding dimensions are set to 64. The learning rate for the ADAM optimizers for $D$ and $G$ are set to 0.0001, with the exponential decay rates for the first and second moments set to 0.5 and 0.999 respectively. 

All our $D$s take the word2vec-transformed vectors as an input and apply two 2-D convolutions, followed by a fully connected layer to return a single value. The dimensions of the second 2-D convolution are the only things varied to address the different input dimensions. Similarly, our $G$s take a random normal noise of size 100 and transform it to the desired output by passing it through a fully-connected layer, and two 2-D fractionally-strided convolution layers. Again, the dimensions of the second fractionally-strided convolution are the only variables to obtain different output dimensions.

Normalizing word vectors after training them has no significant effect on the performance of GAN2vec, and all the results that we present do not carry out this step. Keeping in punctuation helped improve performance, as expected, and none of the experiments filter them out.

To facilitate stable GAN training, we make the following modifications, covered by \cite{chintala2016train}, by running a few preliminary tests on a smaller sample of our dataset:

\begin{itemize}
    \item Use LeakyRELU instead of RELU
    \item Send generated and real mini-batches to $D$ in separate batches
    \item Use label smoothing by setting the target labels to 0.9 and 0.1 instead of 1 and 0 for real and fake samples respectively (for most of our experiments).
\end{itemize}

\section{Metrics}

\subsection{BLEU}
BLEU \citep{papineni2002bleu} originated as a way to measure the quality of machine translation given certain ground truth. Many text generation papers use this as a metric to compare the quality of the generated samples to the target corpus. A higher n-gram coverage will yield a higher BLEU score, with the score reaching a 100\% if all the generated n-grams are present in the corpus. The two potential flaws with this metric are: 1) It does not take into account the diversity of the text generation, this leads to a situation where a mode-collapsing $G$ that produces the same one sentence from the corpus gets a score of 100\%. 2) It penalizes the generation of grammatically coherent sentences with novel n-grams, just because they are absent from the original corpus. Despite these problems, we use BLEU to be consistent with other GANs for text papers. We also present generated samples for the sake of qualitative evaluation by the reader.

\subsection{Self-BLEU}
Self-BLEU is introduced as a metric to measure the diversity of the generated sentences. It does a corpus-level BLEU on a set of generated sentences, and reports the average BLEU as a metric for a given model. A lower self-BLEU implies a higher diversity in the generated sentences, and accordingly a lower chance that the model has mode collapsed. 

It is not clear from \cite{zhu2018texygen}'s work on how many sentences \textit{Texygen} generates to calculate Self-BLEU. For purposes of GAN2vec's results, we produce $1000$ sentences, and for every sentence do a corpus-level BLEU on remaining $999$ sentences. Our results report the average BLEU across all the outputs.

\section{Chinese Poetry Dataset}
\label{sec:chin_results}
The Chinese Poetry dataset, introduced by \citep{zhang-lapata:2014:EMNLP2014} presents simple 4-line poems in Chinese with a length of 5 or 7 tokens (henceforth referred to Poem 5 and Poem 7 respectively). Following previous work by \citep{rajeswar2017adversarial} and \citep{yu2017seqgan}, we treat every line as a separate data point. We modify the Poem 5 dataset to add start and end of tokens, to ensure the model captures (at least) that pattern through the corpus (given our lack of Chinese knowledge). This setup allows us to use identical architectures for both the Poem 5 and Poem 7 datasets. We also modify the GAN2vec loss function with the objective in Eq. \ref{eq:gp}, and report the results below.

\begin{table}[ht]
\centering
\resizebox{0.45\textwidth}{!}{
\begin{tabular}{|p{0.1\linewidth}|p{0.2\linewidth}|p{0.3\linewidth}|p{0.3\linewidth}|}
    \hline
  & \citeauthor{rajeswar2017adversarial} & \textbf{GAN2vec} & \textbf{GAN2vec (wGAN)} \\ \hline
  \textbf{Poem 5}  & -- (train) & 37.90\% (train) & \textbf{53.5\%} (train)\\
  & 87.80\% (test) & 22.50\% (test) & \textbf{25.78\%} (test)\\ \hline
   \textbf{Poem 7}  & -- (train) & 30.14\% (train) & \textbf{66.45\%} (train)\\
  & 65.60\% (test) & 10.20\% (test) & \textbf{22.07\%} (test)\\ \hline

\end{tabular}
}
\caption{Chinese Poetry BLEU-2 scores.}
\label{tab:chin}

\end{table}

The better performance of the GAN2vec model with the wGAN objective is in-line with the image results in \citet{gulrajani2017improved}'s work. We were not able to replicate \cite{rajeswar2017adversarial}'s model on the Chinese Poetry dataset to get the reported results on the test set. This conclusion is in-line with our expectation of lower performance on the test set, given the small overlap in the bi-gram coverage between the provided train and test sets. \cite{lu2018neural} also point out that this work is unreliable, and that their replicated model suffered from severe mode-collapse. On 1000 generated sentences of the Poem-5 dataset, our model has a self BLEU-2 of \textbf{66.08\%} and self BLEU-3 of \textbf{35.29\%}, thereby showing that our model does not mode collapse.

\section{CMU-SE Dataset}
CMU-SE\footnote{\url{https://github.com/clab/sp2016.11-731/tree/master/hw4/data}} is a pre-processed collections of simple English sentences, consisting of 44,016 sentences and a vocabulary of 3,122-word types. For purposes of our experiments here, we limit the number of sentences to 7, chosen empirically to capture a significant share of the examples. For the sake of simplicity in these experiments, for the \textit{real} corpus, sentences with fewer than seven words are ignored, and those with more than seven words are cut-off at the seventh word. 

Table 1 presents sentences generated by the original GAN2vec model. Appendix \ref{sec:gen} includes additional examples. While this is a small subset of randomly sampled examples, on a relatively simple dataset, the text quality appears competitive to the work of \cite{rajeswar2017adversarial} on this corpus.

\begin{table}[ht]
\centering
\begin{tabular}{|l|}
\hline
\citet{rajeswar2017adversarial}\\
\hline
 \textless s\textgreater \, will you have two moment ? \textless/s\textgreater \,              \\
\textless s\textgreater \, how is the another headache ? \textless/s\textgreater \,    \\
\textless s\textgreater \, what ’s in the friday food ? ? \textless/s\textgreater \,      \\
\textless s\textgreater \, i ’d like to fax a newspaper . \textless/s\textgreater \,       \\  \hline
\textbf{GAN2vec} \\
\hline
 \textless s\textgreater \, i dropped my camera . \textless/s\textgreater \, \\
\textless s\textgreater \, i 'd like to transfer it  \\
 \textless s\textgreater \, i 'll take that car ,                         \\
 \textless s\textgreater \, prepare whisky and coffee , please            
  \\  \hline
\end{tabular}
\caption{Example sentences generated by the original GAN2vec. We report example sentences from \citet{rajeswar2017adversarial} and from our GAN2vec model on CMU-SE.}
\label{tab:norm}
\end{table}

\subsection{Conditional GAN2vec}

\label{sec:cond}
We split the CMU-SE dataset into \emph{questions} and \emph{sentences}, checking for the presence of a question mark. We modify the original GAN2vec, as seen in Section \ref{sec:cond_arc}, to now include these labels. Our conditional GANs learn to generate mainly coherent sentences on the CMU-SE dataset, as seen in Table \ref{tab:cond}.

\begin{table*}[!ht]
\centering
\begin{tabular}{|l|l|}
    \hline
  \textbf{Questions} & \textbf{Sentences} \\ \hline
\textless s\textgreater \, can i get you want him                     & \textless s\textgreater \, i bring your sweet inexpensive beer                     \\
\textless s\textgreater \,  where 's the hotel ?           & \textless s\textgreater \, they will stop your ship at              \\
\textless s\textgreater \, what is the fare ? \textless/s\textgreater \, & \textless s\textgreater \, i had a pocket . \textless/s\textgreater \, \\
\textless s\textgreater \, could you buy the timetable ?   & \textless s\textgreater \, it 's ten at detroit western \\ \hline  
\end{tabular}
\caption{Examples of sentences generated by the conditional GAN. We report examples of sentences with our model conditioned on sentence type, i.e., question or sentence.}
\label{tab:cond}
\end{table*}

\begin{figure}[!ht]

\centering
  \begin{minipage}{1.0\columnwidth}
  \centering
    \includegraphics[width=0.8\linewidth]{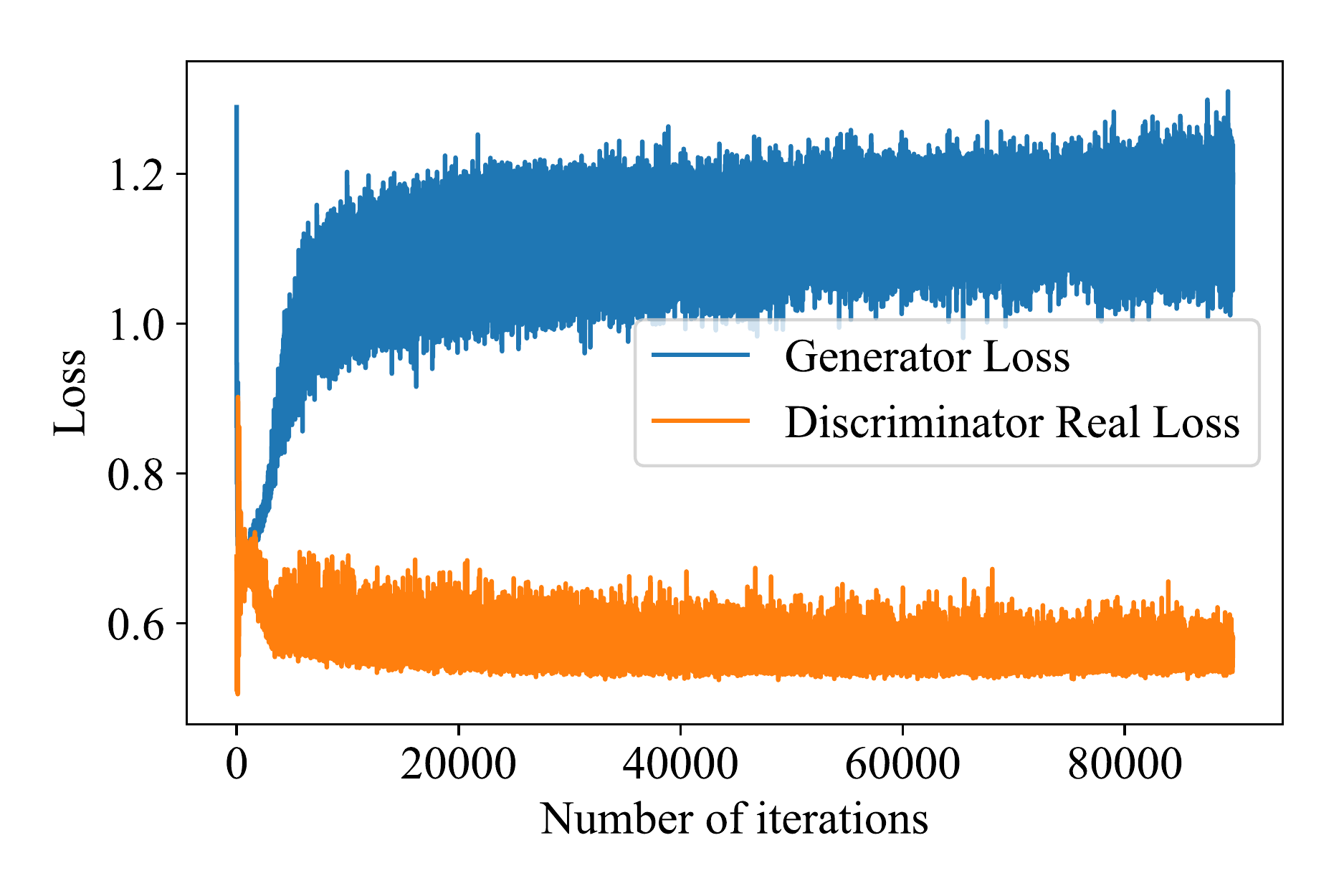}
  \end{minipage}\hfill
  \begin{minipage}{1.0\columnwidth}
  \centering
    \includegraphics[width=0.8\linewidth]{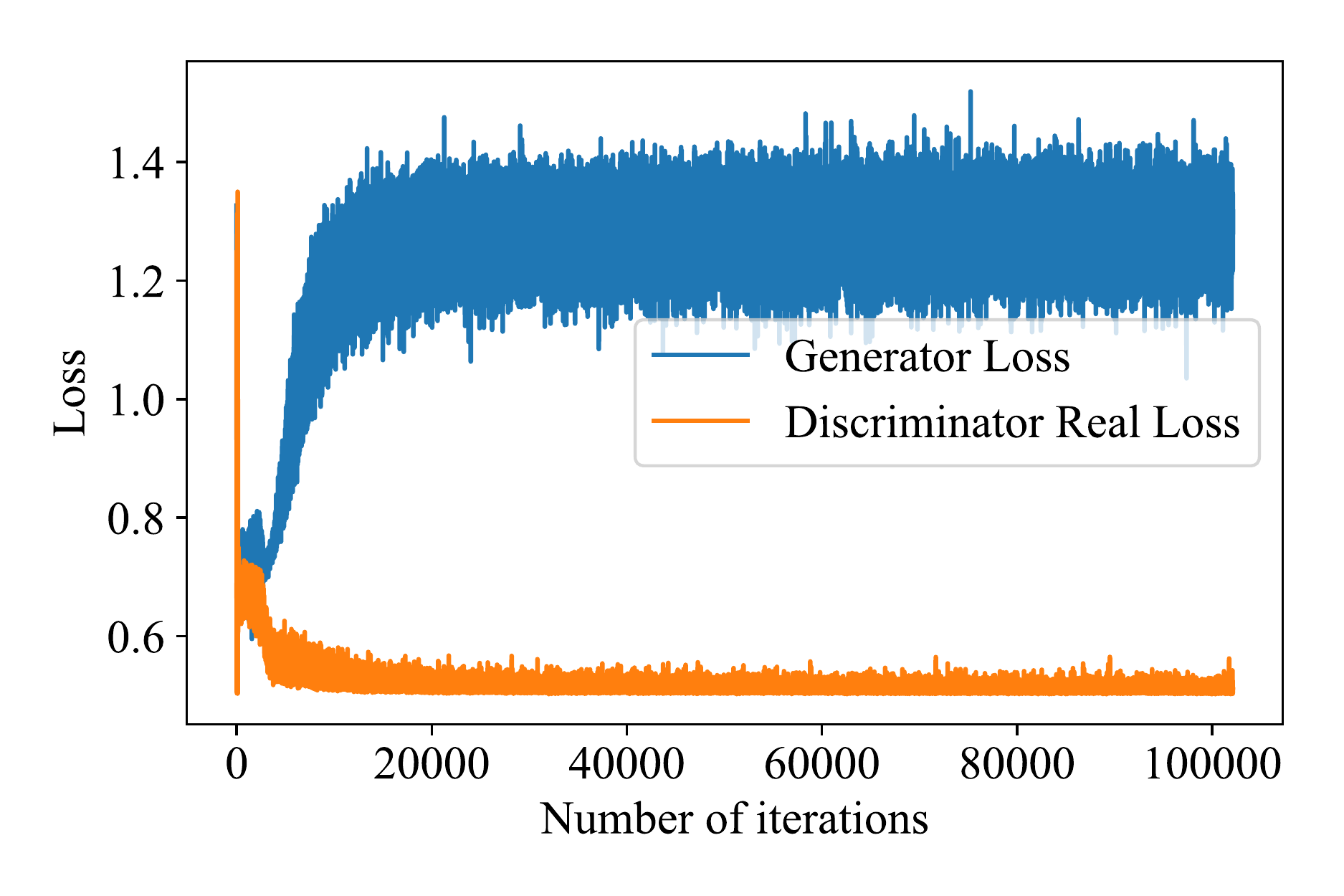}
  \end{minipage}
  \caption{The minimax loss of D and G, with increasing iterations for the GAN2vec model (top) and the conditional GAN2vec (bottom).}
  \label{fig:normgans}
\end{figure}

Figure \ref{fig:normgans} shows the loss graphs for our GAN2vec and conditional GAN2vec trained for $\sim$300 epochs. As seen above, the conditional GAN2vec model generates relatively atypical sentences. This is supported by the second loss curve in Figure \ref{fig:normgans}. The G loss follows a progression similar to the normal GAN2vec case, but the loss is about 16\% more through the 100 epochs.

\subsection{Hyperparameter Variation Study}
We study the effects of different initial hyperparameters for GAN2vec by reporting the results in Table \ref{mlabel}. All the experiments were run ten times, and we report the best scores for every configuration. It must be noted that for conditional GAN2vec training for this experiment, we randomly sample points from the CMU-SE corpus to enforce a 50-50 split across the two labels (question and sentence).

\begin{table*}[!ht]
\centering
\resizebox{0.8\textwidth}{!}{
\begin{tabular}{|l|l|l|l|l|l|l|}
\hline
\textbf{Architecture} & \textbf{Conditional} & \textbf{Vector Type} & \textbf{Loss function} & \textbf{BLEU-2} & \textbf{BLEU-3} \\ \hline
R.1 & No & Sense2vec & Original & 0.743 & 0.41 \\ \hline
R.1 & No & Sense2vec & wgan & 0.7933 & 0.4728 \\ \hline
R.1 & No & Word2Vec & wgan & 0.74 & 0.43 \\ \hline
C.1 & Yes & word2vec & Original (Real) & 0.717 & 0.412 \\ \hline
C.1 & Yes & word2vec & Original & 0.743 & 0.4927 \\ \hline
C.1 & Yes & word2vec & wgan & \textbf{0.7995} & \textbf{0.5168} \\ \hline
C.1 & Yes & word2vec & wgan & \textbf{0.821} & \textbf{0.51} \\ \hline
C.2 & Yes & word2vec & wgan & \textbf{0.8053} & \textbf{0.499} \\ \hline
\end{tabular}
}
\caption{Performance of different models on the CMU-SE train dataset. $R.1$ is the original GAN2vec, $C.1$ is $R.1$ modified with addition of labels, $C.2$ adds batch normalization on the CNN layer of $G$. Original (Real) sets the real label to 0.9, the rest use 1. }
\label{mlabel}
\end{table*}

The overall performance of most of the models is respectable, with all models generating grammatically coherent sentences. GAN2vec with wGAN objective outperforms original GAN2vec, and is inline with the results of \cite{gulrajani2017improved} and our results in Section \ref{sec:chin_results}. Sense2vec does not have a significant improvement over the original word2vec representations. In agreement with \cite{goodfellow2016nips}, providing labels in the conditional variant leads to better performance. 

\subsection{Word2vec cosine similarity}
\begin{figure}[!ht]
\centering
  \includegraphics[width=0.8\linewidth]{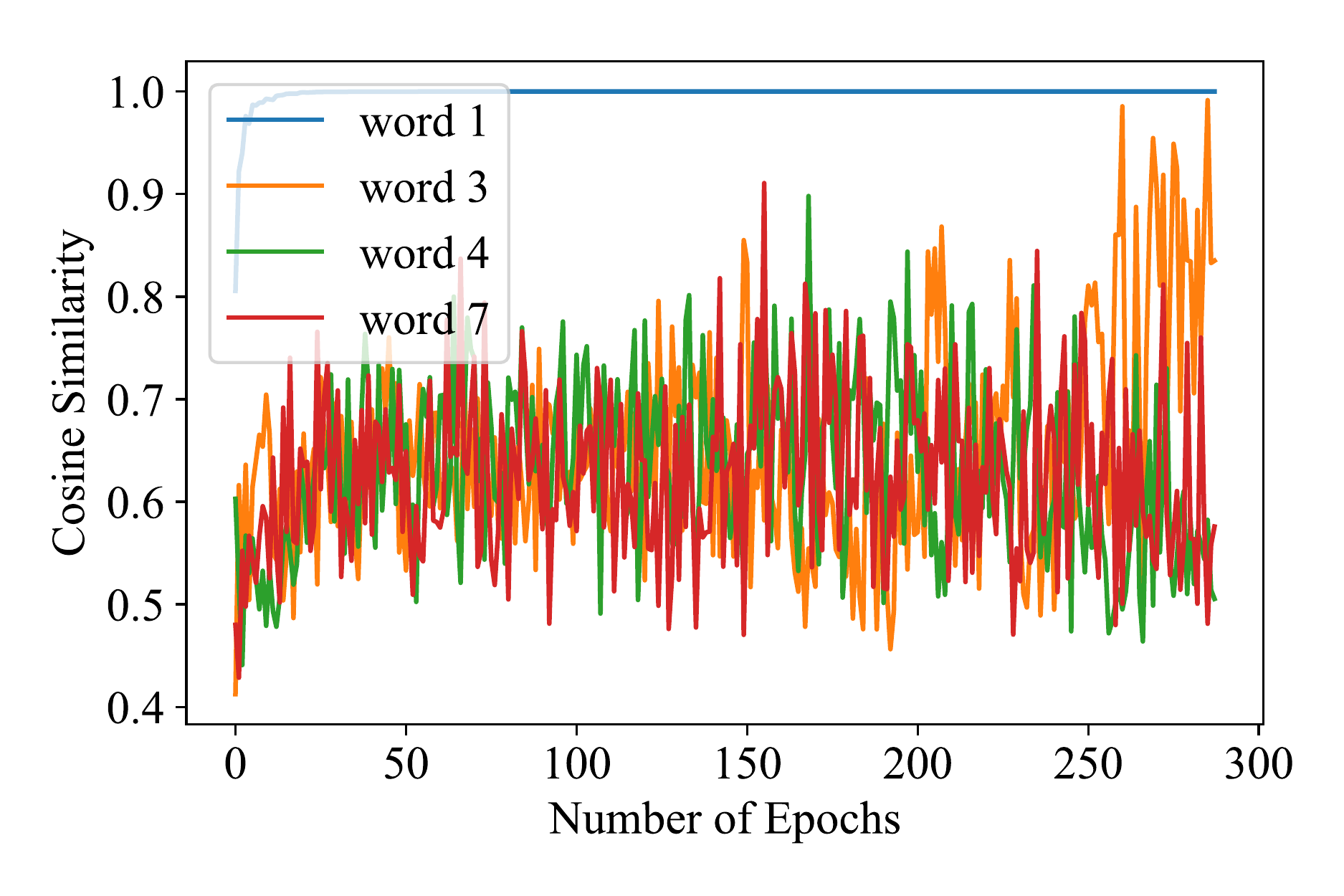}
  \caption{Cosine similarities of the first, third, fourth, and seventh words to the closest words from sentences generated by GAN2vec trained on the CMU-SE dataset.}
  \label{fig:cosine}
\end{figure}
During training, we map our generated word2vec vectors to the closest words in the embedding space and measure the point-wise cosine similarity of the generated vector and the closest neighbour's vector. Figure \ref{fig:cosine} shows these scores for the first, third, fourth and seventh word of the 7-word generated sentences on the CMU-SE dataset for about 300 epochs. The model immediately learns that it needs to start a sentence with \textless s\textgreater \  and gets a cosine similarity of around 1. For the other words in that sentence, the model tends to get better at generating word vectors that are close to their real-valued counterparts of the nearest neighbours. It seems as if the words close to the start of the sentence follow this trend more strongly (as seen with words 1 and 3) and it is relatively weaker for the last word of the sentence.

\section{Coco Image Captions Dataset}
The Coco Dataset is used to train and generate synthetic data as a common dataset for all the best-performing models over the last two years. In \textit{Texygen}, the authors set the sentence length to 20. They train an oracle that generates 20,000 sentences, with one half used as the training set and the rest as the test set. All the models in this benchmark are trained for 180 epochs.

\begin{figure}[!ht]
\centering
  \includegraphics[width=\linewidth]{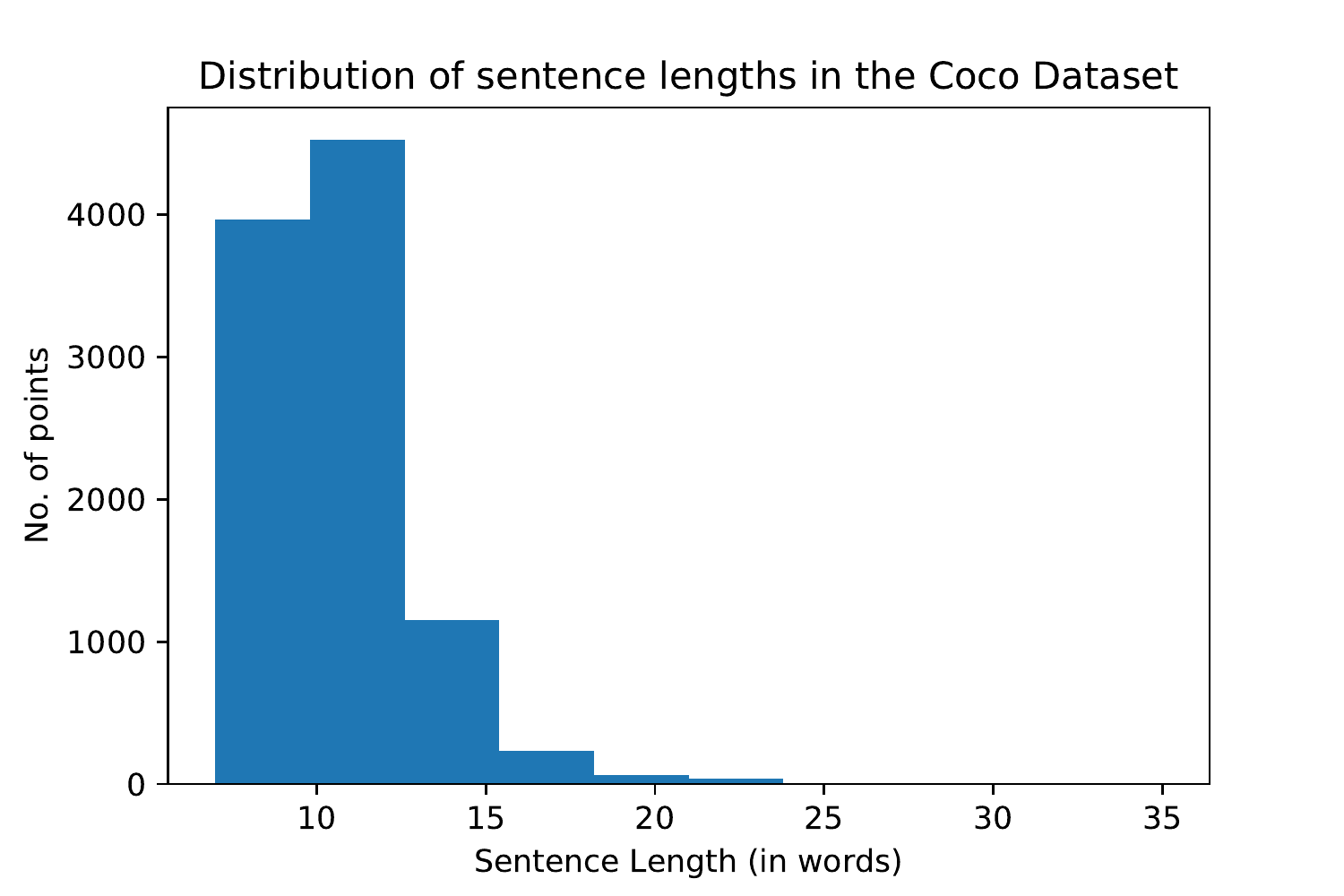}
  \caption{Distribution of sentence lengths in the Coco dataset. Most of the captions have less than 20 words, the cut-off set by \textit{Texygen}.}
  \label{fig:coco}
\end{figure}

Figure \ref{fig:coco} shows the distribution of the sentence lengths in this corpus. For purposes of studying the effects of longer training sentences on GAN2vec, we set the sentence lengths to 7, 10 and 20 (with the respective models labeled as GAN2vec-7, GAN2vec-10, GAN2vec-20 going forward). Any sentence longer than the pre-defined sentence length is cut off to include only the initial words. Sentences shorter than this length are padded with an end of sentence character to fill up the remaining words (we use a comma ($,$) for purposes of our experiments as all the sentences in the corpus end with either a full stop or a word). We tokenize the sentences using NLTK's word tokenizer\footnote{\url{https://www.nltk.org/api/nltk.tokenize.html\#nltk.tokenize.word_tokenize}} which uses regular expressions to tokenize text as in the Penn Treebank corpus\footnote{\url{https://catalog.ldc.upenn.edu/docs/LDC95T7/cl93.html}}. We also report the results of a naive \textit{split at space} approach for the GAN2vec-20 architecture (GAN2vec-20-a), to compare different ways of tokenizing the corpus. We only use the objective from Equation \ref{eq:gp}, given its superior performance to original GAN2vec, as seen in the previous sections.

The results are summarized in the tables below:

\begin{table}[!ht]
\centering
\resizebox{0.4\textwidth}{!}{%
\begin{tabular}{lll}
\hline
\textbf{Model} & \textbf{BLEU-2} & \textbf{BLEU-3} \\ \hline
LeakGAN & 0.926 & 0.816 \\
SeqGAN & 0.917 & 0.747 \\
MLE & 0.731 & 0.497 \\
TextGAN & 0.65 & 0.645 \\
\hline
GAN2vec-7 & 0.548 & 0.271 \\
GAN2vec-10 & 0.641 & \textbf{0.342} \\
GAN2vec-20-a & 0.618 & 0.294\\
GAN2vec-20 & \textbf{0.661} & 0.335 \\
\hline
\end{tabular}
}
\caption{Model BLEU scores on Train Set of the Coco Dataset (higher is better).}
\label{table:train_set}
\end{table}

On the train set (Table \ref{table:train_set}), GAN2vec models have BLEU-2 scores comparable to its SOTA counterparts, with the GAN2vec-20 model having better bigram coverage that TextGAN. The BLEU-3 scores, even though commendable, do not match up as well, possibly signaling that our models cannot keep coherence through longer sentences. The increase in the cut-off sentence length, surprisingly, does not degrade performance. As expected, a trained word tokenizer outperforms its \textit{space-split} counterpart. The performance of the GAN2vec models on the test set (Table \ref{table:test_set}) follows the same trends as that on the train set.
\begin{table}[!ht]
\centering
\resizebox{0.4\textwidth}{!}{%
\begin{tabular}{lll}
\hline
\textbf{Model} & \textbf{BLEU-2} & \textbf{BLEU-3} \\ \hline
LeakGAN & 0.746 & 0.816 \\
SeqGAN & 0.745 & 0.53 \\
MLE & 0.731 & 0.497 \\
TextGAN & 0.593 & 0.645 \\
\hline
GAN2vec-7 & 0.429 & 0.196 \\
GAN2vec-10 & 0.527 & \textbf{0.245} \\
GAN2vec-20-a & 0.484 & 0.206\\
GAN2vec-20 & \textbf{0.551} & 0.232 \\
\hline
\end{tabular}
}
\caption{Model BLEU scores on Test Set of the Coco Dataset (higher is better).}
\label{table:test_set}
\end{table}

\begin{table}[!ht]
\centering
\resizebox{0.4\textwidth}{!}{%
\begin{tabular}{lll}
\hline
\textbf{Model} & \textbf{BLEU-2} & \textbf{BLEU-3} \\ \hline
LeakGAN & 0.966 & 0.913 \\
SeqGAN & 0.95 & 0.84 \\
MLE & 0.916 & 0.769 \\
TextGAN & 0.942 & 0.931 \\
\hline
GAN2vec-7 & \textbf{0.537} & \textbf{0.254} \\
GAN2vec-10 & 0.657 & 0.394 \\
GAN2vec-20-a & 0.709 & 0.394\\
GAN2vec-20 & 0.762 & 0.518 \\
\hline
\end{tabular}
}
\caption{Self BLEU scores of the models trained on the Coco dataset (lower is better).}
\label{table:self_bleu}
\end{table}

Table \ref{table:self_bleu} reports the self-BLEU scores, and all the GAN2vec models significantly outperform the SOTA models, including MLE. This implies that GAN2vec leads to more diverse sentence generations and is less susceptible to mode collapse.

\section{Discussions}

Overall, GAN2vec can generate grammatically coherent sentences, with a good bi-gram and tri-gram coverage from the chosen corpus. BLEU does not reward the generation of semantically and syntactically correct sentences if the associated n-grams are not present in the corpus, and coming up with a new standard evaluation metric is part of on-going work. GAN2vec seems to have comparable, if not better, performance compared to \citet{rajeswar2017adversarial}'s work on two distinct datasets. It depicts the ability to capture the critical nuances when trained on a conditional corpus. While GAN2vec performs slightly worse than most of the SOTA models using the \textit{Texygen} benchmark, it can generate a wide variety of sentences, possibly given the inherent nature of word vectors, and is less susceptible to mode collapse compared to each of the models. GAN2vec provides a simple framework, with almost no overhead, to transfer state of the art GAN research in computer vision to natural language generation. 

We observe that the performance of GAN2vec gets better with an increase in the cut-off length of the sentences. This improvement could be because of \textit{extra} training points for the model. The drop from BLEU-2 to BLEU-3 scores is more extreme than the other SOTA models, indicating that GAN2vec may lack the ability to generate \emph{long} coherent sentences. This behavior could be a manifestation of the chosen $D$ and $G$ architectures, specifically the filter dimensions of the convolution neural networks. Exploration of other structures, including RNN-based models with their ability to remember long term dependencies, might be good alternatives to these initial architecture choices. Throughout all the models in the \textit{Texygen} benchmark, there seems to be a mild negative correlation between diversity and performance. GAN2vec in its original setup leans more towards the generation of new and diverse sentences, and modification of its loss function could allow for tilting the model more towards accurate NLG.

\section{Conclusion}
While various research has extended GANs to operate on discrete data,  most approaches have approximated the gradient in order to keep the model end-to-end differentiable. We instead explore a different approach, and work in the continuous domain using word embedding representations. The performance of our model is encouraging in terms of BLEU scores, and the outputs suggest that it is successfully utilizing the semantic information encoded in the word vectors to produce new, coherent and diverse sentences.
\bibliography{naaclhlt2019}
\bibliographystyle{acl_natbib}

\clearpage
\appendix
\section{CMU-SE}
\label{ap}
\subsection{Conditional Architecture}
\label{sec:cond_arc}
\begin{figure}[ht]
\centering
  \includegraphics[width=\linewidth]{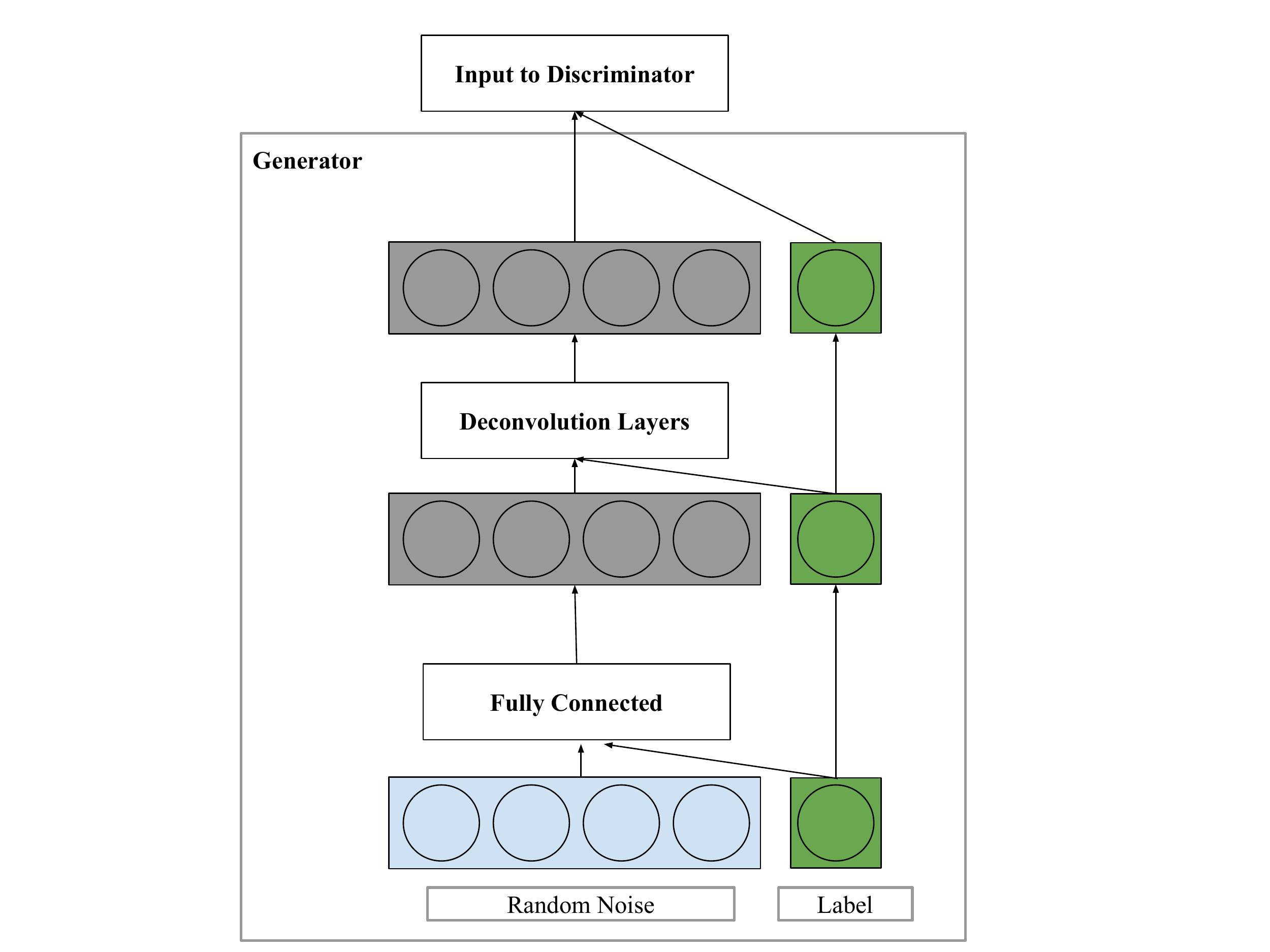}
  \caption{Generator Architecture for Conditional GAN2vec.}
  \label{fig:cond}
\end{figure}

While designing GAN2vec to support conditional labels, as presented in \citet{mirza2014conditional}, we used the architecture in Figure \ref{fig:cond} for our G. The label is sent as an input to both the fully connected and the de-convolution neural layers. The same change is followed while updating D to support document labels.

\label{ap:cond}

\subsection{Examples of Generated Sentences}
\label{sec:gen}

\begin{table*}[ht]
\centering
\caption{Sentences Generated with GAN2vec on CMU-SE}
\label{my-label}
\begin{tabular}{@{}l@{}}
\toprule
\multicolumn{1}{c}{\textbf{Generated Sentences}}                                             \\ \midrule
\textless s\textgreater \, can you have a home \textless/s\textgreater \,                           \\
\textless s\textgreater \, i 'd like to leave the                                                \\
\textless s\textgreater \, this is the baggage . \textless/s\textgreater \,                         \\
\textless s\textgreater \, i 'd like a driver ?                                                  \\
\textless s\textgreater \, do you draw well . \textless/s\textgreater \,                            \\
\textless s\textgreater \, i 'd like to transfer it                                              \\
\textless s\textgreater \, please explain it \textless unk\textgreater . \textless/s\textgreater \,  \\
\textless s\textgreater \, can i book a table .                                                  \\
\textless s\textgreater \, i 'll take that car ,                                                 \\
\textless s\textgreater \, would i like a stay ?                                                 \\
\textless s\textgreater \, will you check it . \textless/s\textgreater \,                           \\
\textless s\textgreater \, do you find it ? \textless/s\textgreater \,                              \\
\textless s\textgreater \, i want some lovely cream .                                            \\
\textless s\textgreater \, could you recommend a hotel with                                      \\
\textless s\textgreater \, can you get this one in                                               \\
\textless s\textgreater \, where 's the petrol station ?                                                 \\
\textless s\textgreater \, what 's the problem ?  \textless/s\textgreater \,                            \\
\textless s\textgreater \, i have a hangover dark .                                              \\
\textless s\textgreater \, i come on the monday .                                                \\
\textless s\textgreater \, i appreciate having a sushi .                                         \\
\textless s\textgreater \, the bus is busy , please                                              \\
\textless s\textgreater \, i dropped my camera . \textless/s\textgreater \,                         \\
\textless s\textgreater \, i want to wash something .                                            \\
\textless s\textgreater \, it 's too great for \textless unk\textgreater                          \\
\textless s\textgreater \, i have a driver in the                                                \\
\textless s\textgreater \, it is delicious and \textless unk\textgreater \textless unk\textgreater \\
\textless s\textgreater \, please leave your luggage . .                                         \\
\textless s\textgreater \, i had alcohol wow , \textless/s\textgreater \,                         \\
\textless s\textgreater \, it is very true .  \textless/s\textgreater \,                         \\
\textless s\textgreater \,  where 's the hotel ?  \textless/s\textgreater \,                         \\
\textless s\textgreater \,   will you see the cat ?   \textless/s\textgreater \,                         \\
\textless s\textgreater \,   where is this bus ?    \textless/s\textgreater \,                         \\
\textless s\textgreater \,  how was the spirits airline warranty                         \\
\textless s\textgreater \,   i would n't sunburn you                          \\

\textless s\textgreater \, prepare whisky and coffee , please                                    \\ \bottomrule
\end{tabular}
\end{table*}

\end{document}